\pdfoutput=1

\documentclass[11pt]{article}

\usepackage[final]{acl}

\usepackage{amsmath,amsfonts,bm}

\usepackage{xspace}
\usepackage{xcolor}
\usepackage{graphicx}
\usepackage{nicematrix}

\newcommand\Comment[1]{}

\newcommand\llm{LLM\xspace}

\newcommand{\ltor}{L2R\xspace}
\newcommand{\fim}{FIM\xspace}
\newcommand{\ltorfull}{Left-to-Right\xspace}
\newcommand{\fimfull}{Fill-in-the-Middle\xspace}
\newcommand{\postp}{post-processing\xspace}
\newcommand{\Postp}{Post-processing\xspace}
\newcommand{\ntpfull}{next-token prediction\xspace}
\newcommand{\ntp}{NTP\xspace}
\newcommand{\oursfull}{horizon-length prediction\xspace} %
\newcommand{\ours}{HLP\xspace} %
\newcommand{\OURSfull}{Horizon-Length Prediction\xspace}

\newcommand{\ourstrain}{NTP + HLP\xspace}

\newcommand{\fimpre}{$\mathsf{prefix}$\xspace}
\newcommand{\fimmid}{$\mathsf{middle}$\xspace}
\newcommand{\fimsuf}{$\mathsf{suffix}$\xspace}

\newcommand{\safim}{SAFIM\xspace}
\newcommand{\cceval}{CrossCodeEval\xspace}
\newcommand{\ccleval}{CrossCodeLongEval\xspace}
\newcommand{\repoeval}{RepoEval\xspace}
\newcommand{\dfourj}{Defects4J\xspace}
\newcommand{\crux}{CRUXEval\xspace}

\newcommand\passat[1]{\mbox{pass@{#1}}}
\newcommand\Passat[1]{\mbox{Pass@{#1}}}

\newcommand{\eg}{\emph{e.g.,}\xspace}
\newcommand{\ie}{\emph{i.e.,}\xspace}

\newcommand{\starcodertwo}{StarCoder2\xspace}

\newcommand\dscoderbase{DeepSeek-Coder-Base\xspace}

\newcommand\humaneval{HumanEval\xspace}
\newcommand\humanevalp{HumanEval+\xspace}
\newcommand\evalplus{EvalPlus\xspace}
\newcommand\mbpp{MBPP\xspace}
\newcommand\mbppp{MBPP+\xspace}

\def\eqref#1{equation~\ref{#1}}

\def\1{\bm{1}}

\DeclareMathAlphabet{\mathsfit}{\encodingdefault}{\sfdefault}{m}{sl}
\SetMathAlphabet{\mathsfit}{bold}{\encodingdefault}{\sfdefault}{bx}{n}

\usepackage{booktabs} %
\usepackage{hyperref}

\usepackage{amsmath}
\usepackage{amssymb}
\usepackage{mathtools}
\usepackage{amsthm}

\usepackage[capitalize,noabbrev]{cleveref}

\usepackage{multirow}
\usepackage{caption}
\usepackage{subcaption}
\usepackage{wrapfig,lipsum}
\usepackage{arydshln}

\usepackage[
  moderate
]{savetrees}
\usepackage{blindtext}

\usepackage{times}
\usepackage{latexsym}

\usepackage[T1]{fontenc}

\usepackage[utf8]{inputenc}

\usepackage{microtype}

\usepackage{inconsolata}

\usepackage{graphicx}

\usepackage{enumitem}
\usepackage[table-number-alignment=center, group-separator={}]{siunitx}
\def\sym#1{\ifmmode^{#1}\else\(^{#1}\)\fi}
\usepackage{array}

\usepackage{etoolbox}
\newrobustcmd\B{\DeclareFontSeriesDefault[rm]{bf}{b}\bfseries}  

\theoremstyle{plain}

\theoremstyle{definition}

\theoremstyle{remark}

\usepackage[textsize=tiny]{todonotes}

\title{Planning-Aware Code Infilling via Horizon-Length Prediction}

\author{
\noindent\begin{tabular}{@{}p{.16\textwidth}@{}p{.172\textwidth}@{}p{.158\textwidth}@{}p{.123\textwidth}@{}p{.175\textwidth}@{}p{.162\textwidth}@{}}
Yifeng Ding$^{1}$\thanks{Work done at AWS AI Labs.} & Hantian Ding$^2$ & Shiqi Wang$^{3*}$ & Qing Sun$^2$ & Varun Kumar$^2$ & Zijian Wang$^{3*}$ 
\end{tabular}\\
$^1$University of Illinois Urbana-Champaign \quad $^2$AWS AI Labs \quad $^3$Meta \\
\texttt{yifeng6@illinois.edu} \quad \texttt{\{dhantian,qinsun,kuvrun\}@amazon.com} \\ \texttt{\{tcwangshiqi,zijianwang\}@meta.com} 
}

\begin{document}
\maketitle

\begin{abstract}

\fimfull (\fim), or infilling, has become integral to code language models, enabling generation of missing code given both left and right contexts. However, the current \fim training paradigm which performs next-token prediction (\ntp) over reordered sequence often leads to models struggling to generate content that aligns well with the surrounding context.
We hypothesize that \ntp alone is insufficient for models to \textit{learn effective planning} conditioned on the distant right context, a critical factor for successful code infilling. To overcome this, we propose \textbf{\OURSfull} (\textbf{\ours}), a novel training objective that teaches models to predict the number of remaining middle tokens at each step. \ours advances \fim with lookahead planning, enabling models to inherently learn infilling boundaries for arbitrary left and right contexts without relying on dataset-specific post-processing. Our evaluation across different model families and sizes shows that \ours significantly improves \fim performance by up to 24\% relatively on diverse benchmarks, across file-level and repository-level. Furthermore, the enhanced planning capability gained through \ours boosts model performance on code reasoning. Importantly, \ours incurs negligible training overhead and no additional inference cost, ensuring its practicality for real-world scenarios.
\end{abstract}

\section{Introduction}
Fill-in-the-Middle (\fim), or infilling, has become essential for modern code development, where programmers frequently need to insert or modify code between existing sections rather than writing linearly from start to end \citep{openaifim, incoder}. While large language models have shown remarkable capabilities in code generation, the \fim task poses unique challenges that go beyond traditional left-to-right generation. A model must not only generate code that follows from the preceding context (\fimpre), but also smoothly connect to the subsequent code (\fimsuf) – a task that requires careful planning and foresight.

\begin{figure}[!tbp]
\centering
\includegraphics[width=1.0\linewidth]{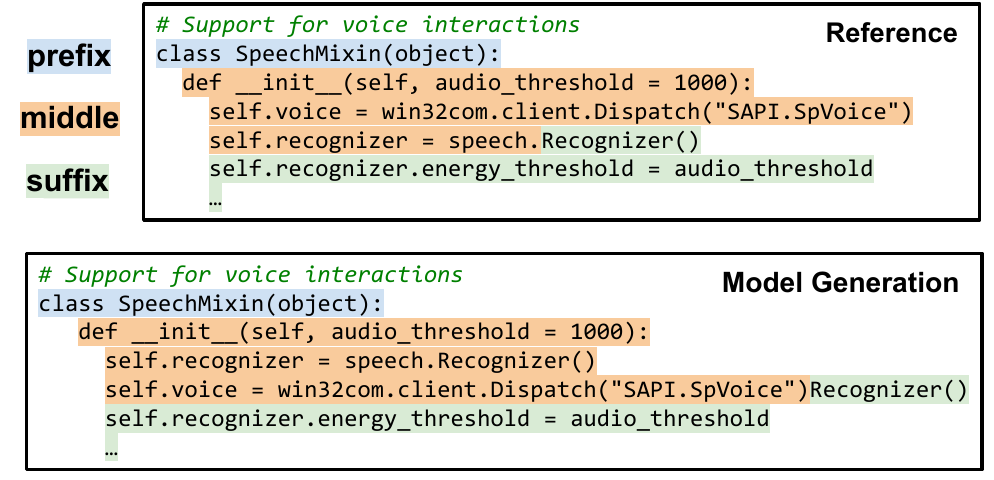}
\caption{Successful \fim requires planning capabilities. Given prefix and suffix, the model is asked to infill the middle part. Compared with the ground truth, \llm fails to connect to \fimsuf due to lack of planning capability: the last part of the generation needs to connect with the member function \texttt{Recognizer()}.
}
\vspace*{-1.6\baselineskip}
\label{fig:fim_plan_example}
\end{figure}

Current approaches to \fim typically reorder the input sequence and apply standard next-token prediction (\ntp) training \citep{starcodertwo, dscoder, dscodervtwo, qwentwofive}. However, as illustrated in Figure \ref{fig:fim_plan_example}, this methodology has a fundamental limitation: models struggle to maintain coherence over longer sequences and often fail to create smooth transitions to the right context \citep{openaifim}. While existing benchmarks attempt to address this through rule-based post-processing (e.g., truncating generated code based on line count \citep{repoeval, repoformer} or program structure \citep{crosscodeeval, safim}), such methods rely on dataset-specific assumptions that do not generalize to real-world scenarios where both left and right contexts can be arbitrary.

\begin{table*}[h]
\centering
\resizebox{2\columnwidth}{!}{
\begin{tabular}{@{}l|l@{}}
\toprule
        & \multicolumn{1}{c}{Post-processing Criteria}                                                                                                          \\ \midrule
\repoeval\citep{repoeval} & \multirow{2}{*}{Truncate generation to \textbf{the same number of lines as in ground truth}.}                                      \\
\ccleval\citep{repoformer} &                                                                  \\ \midrule
\cceval\citep{crosscodeeval}  & Truncate generation \textbf{at the first complete statement}.                                                                   \\ \midrule
\safim\citep{safim}   & Stop when \textbf{the target program structure in ground truth} is generated. \\ \bottomrule
\end{tabular}
}
\caption{\label{tab:unrealistic}
Post-processing criteria used in existing \fim benchmarks. Text in bold denotes restrictive dataset-specific knowledge they employ in evaluation.
}
\vspace*{-\baselineskip}
\end{table*}

We hypothesize that the core challenge lies in the model's limited ability to plan ahead while filling in the middle. Standard next-token prediction training only requires considering one token at a time, but successful code infilling demands awareness of the entire missing section to ensure both local coherence and proper connection to the right context. This planning capability is particularly crucial because generated code must satisfy both syntactic requirements and semantic constraints from both surrounding contexts.

To address this limitation, we propose Horizon-Length Prediction (\ours), a novel training objective that teaches models to predict the number of remaining tokens needed to complete the middle section at each generation step. \ours advances \ntp by encouraging models to develop awareness of the generation horizon and plan accordingly. Unlike post-processing approaches, \ours is generalizable as it does not rely on any task-specific knowledge. Instead, it strengthens models' innate ability to plan and execute coherent code completions.

Our comprehensive evaluation across different models and model sizes demonstrates that HLP significantly improves \fim performance, achieving up to 24\% relative improvement on diverse benchmarks at both file-level and repository-level without relying on  any dataset-specific post-processing. Furthermore, the enhanced planning capability gained through HLP training also boosts model performance on code reasoning tasks. Importantly, HLP incurs negligible training overhead and no additional inference cost, making it practical for real-world deployment.

Our key contributions are as follows:

\vspace{-\topsep}

\begin{itemize}[leftmargin=1em]
    \setlength{\parskip}{2pt}
    \item We identify planning capability as a fundamental bottleneck in current FIM approaches and quantitatively demonstrate how post-processing methods in existing benchmarks obscure this critical limitation.
    \item We propose  Horizon-Length Prediction (HLP), a novel training objective that advances fill-in-the-middle capability by teaching models to plan over arbitrarily long horizons,  with negligible training and zero inference overhead.

    \item We demonstrate up to 24\% improvement in FIM performance across multiple benchmarks and model families without relying on any post-processing.
    \item We show that HLP's benefits extend beyond FIM to improve performance on code reasoning tasks, and present various analyses that illuminate the underlying mechanism that enable these improvements.

\end{itemize}

\section{\Postp for \fimfull}\label{sec:postp}

Most existing \fim works rely on \postp to truncate code completions generated by \llm{s} for infilling tasks \citep{safim, repoeval, crosscodeeval, repoformer}. While such \postp can enhance the \fim performance, we argue that they fundamentally depend on dataset-specific assumptions that make them impractical for real-world scenarios (\S\ref{sec:post_unrealistic}). Through evaluation, we show that the performance of \fim existing code models drops significantly without \postp, suggesting that \postp conceals the fundamental struggles of models with code filling (\S\ref{sec:post_obfuscate}). Furthermore, we show that this limitation stems from models' inability to plan coherent completions given a fixed \fimsuf – a challenge that post-processing alone cannot address.

\subsection{\Postp Requires Task-Specific Knowledge}\label{sec:post_unrealistic}

\Postp methods adopted by recent \fim benchmarks typically assume a certain completion type and perform rule-based truncation accordingly. Table \ref{tab:unrealistic} summarizes the \postp criteria of four popular \fim benchmarks, highlighting the specific rule used for each dataset. These criteria do not transfer across datasets, nor are they generalizable to \fim in the real-world scenario where both left and right contexts can be arbitrary. Given the complexity of programming languages, developing universally applicable post-processing rules for infilling is infeasible. Instead, models need to learn the intrinsic patterns that make for good completions.

\begin{table}[t]
\centering
\resizebox{0.92\columnwidth}{!}{
\begin{tabular}{@{}lccccc@{}}
\toprule
                    & \multicolumn{4}{c}{SAFIM}                                                                                                    & \multicolumn{1}{c}{\multirow{2}{*}{Avg}} \\ \cmidrule(r){2-5}
                    & \multicolumn{1}{c}{Algo} & \multicolumn{1}{c}{Algo$_{\text{v2}}$} & \multicolumn{1}{c}{Control} & \multicolumn{1}{c}{API} & \multicolumn{1}{c}{}                         \\ \midrule
DS-1.3B             & \multicolumn{1}{l}{}            & \multicolumn{1}{l}{}               & \multicolumn{1}{l}{}        & \multicolumn{1}{l}{}    & \multicolumn{1}{l}{}                         \\
\ \ w/ post  & 43.9                   & 49.2                      & 55.6               & 62.9           & 52.9                                \\ 
\ \ w/o post & 39.8                            & 42.4                               & 52.4                        & 56.1                    & 47.7                                         \\
\addlinespace[0.1em]
\hdashline
\addlinespace[0.3em]
\ \ rel. diff & \textcolor{red}{\textbf{-9.3\%}}                            & \textcolor{red}{\textbf{-13.8\%}}                               & \textcolor{red}{\textbf{-5.8\%}}                        & \textcolor{red}{\textbf{-10.8\%}}                    & \textcolor{red}{\textbf{-9.9\%}}                                         \\ \midrule
DS-6.7B             & \multicolumn{1}{l}{}            & \multicolumn{1}{l}{}               & \multicolumn{1}{l}{}        & \multicolumn{1}{l}{}    & \multicolumn{1}{l}{}                         \\ 
\ \ w/ post  & 54.9                   & 58.9                      & 68.1               & 71.0           & 63.2                                \\ 
\ \ w/o post & 53.4                            & 56.7                               & 66.6                        & 69.0                    & 61.4                                         \\
\addlinespace[0.1em]
\hdashline
\addlinespace[0.3em]
\ \ rel. diff & \textcolor{red}{\textbf{-2.7\%}}                            & \textcolor{red}{\textbf{-3.7\%}}                               & \textcolor{red}{\textbf{-2.2\%}}                        & \textcolor{red}{\textbf{-2.8\%}}                    & \textcolor{red}{\textbf{-2.8\%}}                                         \\ \midrule
SC2-3B              &                                 &                                    &                             &                         &                                              \\ 
\ \ w/ post  & 48.1                   & 53.5                      & 60.1               & 68.4           & 57.5                                \\ 
\ \ w/o post & 45.4                            & 49.7                               & 57.1                        & 61.3                    & 53.4                                         \\
\addlinespace[0.1em]
\hdashline
\addlinespace[0.3em]
\ \ rel. diff & \textcolor{red}{\textbf{-5.6\%}}                            & \textcolor{red}{\textbf{-7.1\%}}                               & \textcolor{red}{\textbf{-5.0\%}}                        & \textcolor{red}{\textbf{-10.4\%}}                    & \textcolor{red}{\textbf{-7.2\%}}                                         \\ \midrule
SC2-7B              &                                 &                                    &                             &                         &                                              \\
\ \ w/ post  & 50.4                   & 55.8                      & 62.3               & 70.3           & 59.7                                \\ 
\ \ w/o post & 48.4                            & 53.1                               & 60.4                        & 63.9                    & 56.5                                         \\
\addlinespace[0.1em]
\hdashline
\addlinespace[0.3em]
\ \ rel. diff & \textcolor{red}{\textbf{-4.0\%}}                            & \textcolor{red}{\textbf{-4.8\%}}                               & \textcolor{red}{\textbf{-3.0\%}}                        & \textcolor{red}{\textbf{-9.1\%}}                    & \textcolor{red}{\textbf{-5.4\%}}                                         \\ \bottomrule
\end{tabular}
}
\caption{\label{tab:postprocess}
Effect of post-processing techniques for different code \llm{s} on \safim, where ``\textbf{w/ post}'' refers to using post-processing, ``\textbf{w/o post}'' refers to not using post-processing, and ``\textbf{rel. diff}'' refers to the relative performance difference between the two. We follow the same settings used in  \S\ref{eval:safim}.
}
\end{table}

\subsection{\llm{s} Fail to Plan Coherent Completions}\label{sec:post_obfuscate}

To further demonstrate to what extent \postp conceals \llm{s}' inability of connecting to \fimsuf, we conduct a comprehensive experiment on \safim. We compare \fim performance of four different code \llm{s}, with or without post-processing. As shown in Table \ref{tab:postprocess}, removing \postp leads to up to a substantial 13.8\% Pass@1 drop across all models. This reveals that current models have not truly mastered the fundamental task of generating code that properly connects \fimpre and \fimsuf contexts. Post-processing creates an illusion of competence by artificially ``fixing'' problematic generations rather than addressing the core limitation.

\subsection{\fim Requires Planning Capability}\label{sec:post_plan_ahead}

The challenges in \fim stem from models' inability to plan coherent completions. Consider the example in Figure \ref{fig:fim_plan_example}, where a model needs to implement a speech recognition initialization: while the model demonstrates knowledge of the required components (using \texttt{Recognizer}), it fails to order them properly. Without careful planning, it prematurely places the \texttt{Recognizer} call, leading to both syntactic and semantic errors. Importantly, this type of failure cannot be fixed through post-processing, as truncating the generation would lose essential statements while keeping it results in invalid code.

Such an example illustrates that successful \fim requires not only knowledge of individual components but also the model's ability to plan coherent sequences that smoothly connect to both contexts. The model must reason about the entire completion considering both local coherence and global structure, and that points to a clear need for models to develop intrinsic planning capabilities to succeed at \fim.

\section{\OURSfull }\label{sec:ours_design}
Given a document $D = \{ x_t \}_{t=1}^{T} $ that contains $T$ tokens $x_1,\ x_2,\cdots,\ x_T$, existing \fim training scheme can be divided into three steps: (1) Split the document $D$ into three parts: \fimpre-\fimmid-\fimsuf\footnote{We opt to use PSM setting in this work given our base models DeepSeek-Coder and StarCoder2 were both pre-trained with PSM. We expect that our method is generalizable to SPM setting as well.}, (2) Construct a new \fim-style document $D'$ by reordering the three parts as \fimpre-\fimsuf-\fimmid, and (3) Conduct \ntpfull (\ntp) training on the document $D'$. 

Specifically, we define the three parts in document $D$ as $\text{\fimpre}=x_{1\cdots P}$, $\text{\fimmid}=x_{P+1\cdots P+M}$, and $\text{\fimsuf}=x_{P+M+1\cdots T}$. Then, the new document $D'$ will be formatted as follows:
\vspace{2mm}
\begin{equation}\label{formula:fim_data}
\begin{split}
&D' = {\small \text{<}\textsc{pre}\text{>}}\ \text{\fimpre}\ {\small \text{<}\textsc{suf}\text{>}}\ \text{\fimsuf}\ {\small \text{<}\textsc{mid}\text{>}}\ \text{\fimmid}\ {\small \text{<}\textsc{eoi}\text{>}} \\
&= {\small \text{<}\textsc{pre}\text{>}}\ x_{1\cdots P}\ {\small \text{<}\textsc{suf}\text{>}}\ x_{P+M+1\cdots T}\ {\small \text{<}\textsc{mid}\text{>}} \\
&\ \ \ \ \ \ \ \ \ \ \ \ \ \ \ \ \ \ \ \ \ \ \ \ \ \ \ \ \ \ \ \ \ \ \ \ \ \ \ \ \ \ \ \ \ \ \ \ \ x_{P+1\cdots P+M}\ {\small \text{<}\textsc{eoi}\text{>}} \\
&\overset{\Delta}{=} y_{1\cdots T-M+3}\ x_{P+1\cdots P+M}\ {\small \text{<}\textsc{eoi}\text{>}},
\end{split}
\end{equation}
where the last step re-indexes the leading tokens up until \text{<}\textsc{mid}\text{>} to $y_{1\cdots T-M+3}$ to focus on the \fim part, as \llm{s} are expected to start infilling after <\textsc{mid}> token and to end generation with <\textsc{eoi}> token to connect to \fimsuf accurately. 

Next-token prediction (\ntp) training is conducted on the document $D'$, which aims to minimize the following cross-entropy loss (where $P_{\theta}$ refers to the \llm{s} being trained):

\begin{equation}
\begin{split}
&L_{\ntp} =  - \sum_{t=1}^{T-M+2}\log P_{\theta}(y_{t+1}|y_{1\cdots t}) \\
& - \sum_{t=1}^{M-1}\log P_{\theta}(x_{P+t+1}|y_{1\cdots T-M+3},x_{P+1\cdots P+t}) \\
& - \log P_{\theta}(\text{<}\textsc{eoi}\text{>}|y_{1\cdots T-M+3},x_{P+1\cdots P+M}).
\end{split}
\end{equation}

\vspace{2mm}

While \ntp provides basic \fim capabilities, it has a fundamental limitation: the model only learns to predict one token at a time without developing awareness of the overall horizon. This makes it difficult to plan coherent sequences that properly connect to both contexts.

\begin{figure*}[h]
\centering
\includegraphics[width=0.8\linewidth]{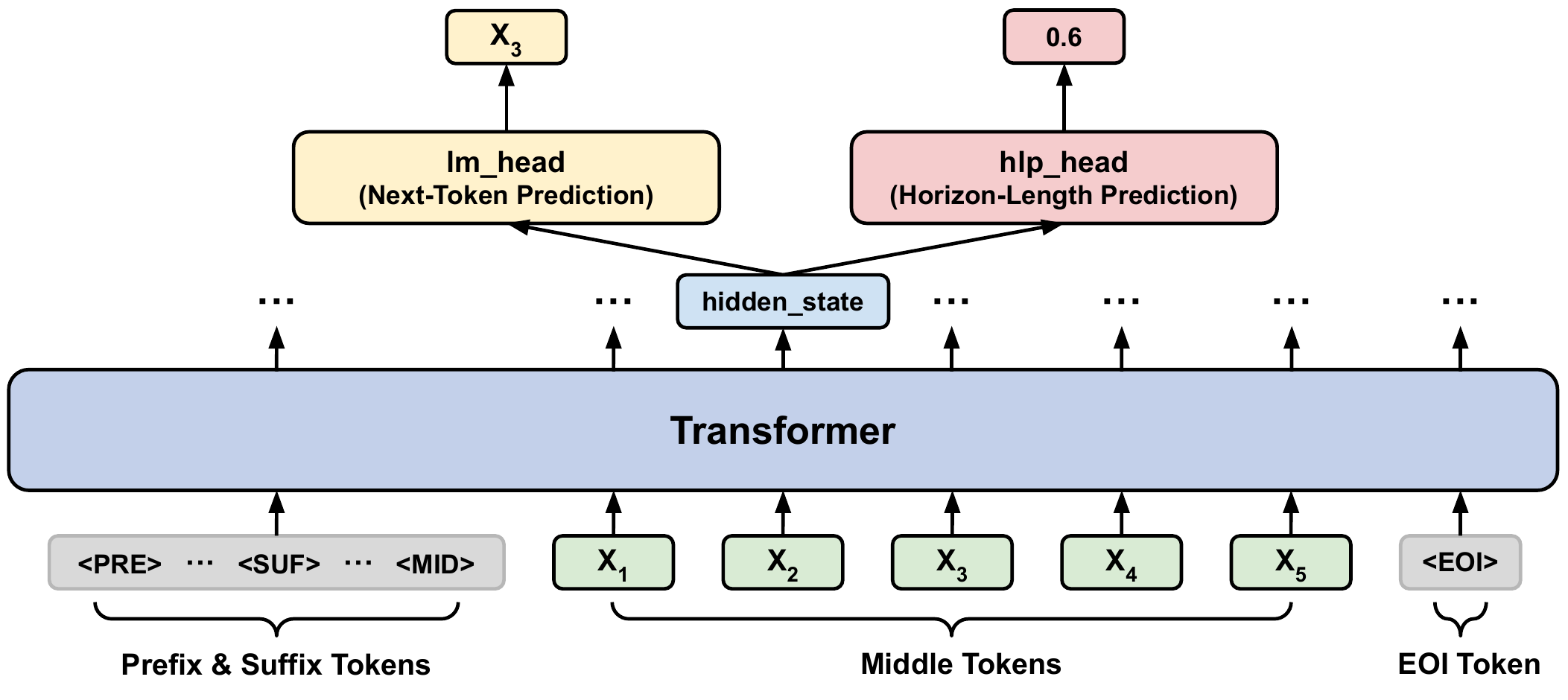}
\caption{Overview of \OURSfull (\ours). In this example, we set the length of \fimmid to five tokens. Following the flow of arrows, we illustrate how the second token of \fimmid (\ie ``$x_2$'') is processed through both next-token prediction objective and horizon-length prediction objective.
}
\label{fig:overview}
\end{figure*}

\textbf{\OURSfull (\ours)}. To mitigate this issue, we propose \oursfull (\ours) as an auxiliary training objective. As shown in Figure  \ref{fig:overview}, the key idea is to teach models to predict the number of remaining tokens needed to complete the \fimmid section at each generation step. This creates an explicit training signal for planning awareness. 

Specifically, at each position $t$ in \fimmid, \ours predicts 
\vspace{1mm}
\begin{equation}\label{formula:ftc_label}
\begin{split}
y_t = \frac{M-t}{M} \in (0, 1].
\end{split}
\end{equation}\vspace{0mm}

where $M$ is the total length of \fimmid. This normalized value represents the portion of \fimmid that remains to be generated and ensures the target is always within $(0, 1]$ interval regardless of the model's context window size. 

\ours is implemented as a linear layer on top of the transformer model (\ie $\mathsf{hlp\_head}$ in Figure~\ref{fig:overview}) with weight $w_{hlp}$, whose input is the hidden state $h_t$ from the last attention layer. The output $w_{hlp}^\top h_t$ is converted to a value between $0$ and $1$ through a sigmoid layer $\sigma$ to represent the final prediction. We use L1 loss for \ours:

\vspace{1mm}

\begin{equation}\label{formula:ftc_loss}
\begin{split}
L_{\ours} = \sum_{t=1}^{M}|\sigma(w_{hlp}^\top h_t) - y_t|.
\end{split}
\end{equation}

\vspace{1mm}

The full training objective is a weighted sum of \ntp loss and \ours loss:
\vspace{-2mm}
\begin{equation}\label{formula:full_loss}
\begin{split}
L = L_{\ntp} + \lambda \cdot L_{\ours},
\end{split}
\end{equation}
where $\lambda$ is the tunable weight. In experiments, we set $\lambda=0.1$, which achieves good results across benchmarks empirically. We further study the relationship between the values of $\lambda$ and the \fim performance of models trained with \ours. As discussed in Appendix \ref{sec:lambdastudy}, the performance of \ours is robust to different values of $\lambda$.

\textbf{Overhead Analysis.} While \ours introduces the additional $\mathsf{hlp\_head}$ during training, the number of added parameters is <\textup{ }0.01\% of the base model, which incurs \textit{almost zero training overhead}. Furthermore, the additional head will not be used during inference, leading to \textit{zero inference overhead}.

\section{Experiments}\label{sec:setup}

\begin{table}[b]
\centering
\sisetup{table-space-text-post = \sym{*}}
\sisetup{table-number-alignment=center}
\resizebox{0.95\columnwidth}{!}{
\begin{tabular}{@{}lccccS[table-align-text-post=false]@{}}
\toprule
                & \multicolumn{4}{c}{SAFIM}                                                                                                    & \multicolumn{1}{c}{\multirow{2}{*}{Average}} \\ \cmidrule(r){2-5}
                & \multicolumn{1}{c}{Algo} & \multicolumn{1}{c}{Algo$_{\text{v2}}$} & \multicolumn{1}{c}{Control} & \multicolumn{1}{c}{API} & \multicolumn{1}{c}{}                         \\ \midrule
DS-1.3B    & 39.8                            & 42.4                               & 52.4                        & 56.1                     & 47.7                                         \\
\addlinespace[0.1em]
\hdashline
\addlinespace[0.3em]
\ \ \ + \ours & \textbf{41.3}                   & \textbf{46.1}                      & \textbf{53.4}               & \textbf{59.0}            & \textbf{50.0\sym{*}}                               \\ \midrule
DS-6.7B   & 53.4        & 56.7               & 66.6          & 69.0                     & 61.4                                         \\
\addlinespace[0.1em]
\hdashline
\addlinespace[0.3em]
\ \ \ + \ours & \textbf{53.5} & \textbf{57.4}      & \textbf{66.9} & \textbf{69.7}            & \textbf{61.9\sym{*}}                                \\ \midrule
SC2-3B      & 45.4                            & 49.7                               & 57.1                        & 61.3                     & 53.4                                         \\
\addlinespace[0.1em]
\hdashline
\addlinespace[0.3em]
\ \ \ + \ours  & \textbf{47.2}                   & \textbf{52.1}                      & \textbf{58.7}               & \textbf{64.5}            & \textbf{55.6\sym{*}}                                \\ \midrule
SC2-7B      & 48.4                            & 53.1                               & 60.4                        & 63.9                     & 56.5                                         \\
\addlinespace[0.1em]
\hdashline
\addlinespace[0.3em]
\ \ \ + \ours  & \textbf{49.4}                   & \textbf{54.5}                      & \textbf{61.8}               & \textbf{65.8}            & \textbf{57.9\sym{*}}                                \\ \bottomrule
\end{tabular}
}
\caption{\label{tab:safim}
\Passat1 results of training w/o and w/ \ours for different code \llm{s} on SAFIM \citep{safim} computed with greedy decoding. We perform statistical significance tests on ``Average'' and show that all results are significant. The same notation applies hereafter.
}
\end{table}

\textbf{Training.} We conduct continual pre-training on a set of code \llm{s} of different model families and sizes to validate the effectiveness of \ours. Specifically, \dscoderbase 1.3B/6.7B \citep{dscoder} and \starcodertwo 3B/7B \citep{starcodertwo} are involved in our experiments. We use AdamW \citep{adamw} as the optimizer with $\beta_1=0.9$ and $\beta_2=0.95$.  We use a cosine learning rate scheduler with a peak learning rate equal to that at the pre-training end. All models are trained for 200K steps with a global batch size of 512.

\textbf{Dataset.} We use a subset of the-stack-v2 \citep{starcodertwo} including Python, Java, C++, and C\#. Following existing works \citep{dscoder, starcodertwo}, \fim rate is set to $0.5$. We employ Best-fit Packing \citep{ding2024fewertruncationsimprovelanguage} to group multiple files into each training sequence while masking out cross-file attention. The \fimpre-\fimmid-\fimsuf split is applied to each file independently rather than the whole training sequence. 

\begin{table*}[h]
\vspace*{-0.4\baselineskip}
\centering
\sisetup{table-space-text-post = \sym{*}}
\sisetup{table-number-alignment=center}
\begin{tabular}{@{}lccccccS[table-format=2.3,table-align-text-post=false]S[table-format=2.3,table-align-text-post=false]@{}}
\toprule
                 & \multicolumn{6}{c}{\cceval\ / \ccleval}                                                                                                                   & \multicolumn{2}{c}{\multirow{2}{*}{Average}}    \\ \cmidrule(r){2-7}
                 & \multicolumn{2}{c}{Line}                        & \multicolumn{2}{c}{Chunk}                       & \multicolumn{2}{c}{Function}                    & \multicolumn{2}{c}{}                            \\ \midrule
                 & \multicolumn{1}{c}{EM} & \multicolumn{1}{c}{ES} & \multicolumn{1}{c}{EM} & \multicolumn{1}{c}{ES} & \multicolumn{1}{c}{EM} & \multicolumn{1}{c}{ES} & \multicolumn{1}{c}{EM\ \ \ \tiny{\ }} & \multicolumn{1}{c}{ES} \\ \midrule
DS-1.3B & 15.23                  & 49.64                  & 22.48                  & 56.40                  & 4.58                   & 33.96                   & 14.10                  & 46.67                  \\
\addlinespace[0.1em]
\hdashline
\addlinespace[0.3em]
\ \ \ + \ours     & \textbf{18.99}         & \textbf{55.47}         & \textbf{24.32}         & \textbf{58.77}         & \textbf{5.12}          & \textbf{35.25}          & \textbf{16.14\sym{*}}         & \textbf{49.83\sym{*}}         \\ \midrule
DS-6.7B & 26.23                  & 62.07                  & 28.90                  & 62.37                  &         \textbf{7.50}           &        \textbf{41.42}            &       20.88            &      55.29             \\
\addlinespace[0.1em]
\hdashline
\addlinespace[0.3em]
\ \ \ + \ours    & \textbf{27.35}         & \textbf{63.54}         & \textbf{30.08}         & \textbf{63.18}         & 7.22          & 40.99          & \textbf{21.55\sym{*}}         & \textbf{55.90\sym{*}}         \\ \midrule
SC2-3B  & 24.17                  & 59.89                  & 22.20                  & 52.69                  & 6.80                   & 38.13                   & 17.72                  & 50.24                  \\
\addlinespace[0.1em]
\hdashline
\addlinespace[0.3em]
\ \ \ + \ours      & \textbf{25.67}         & \textbf{62.62}         & \textbf{30.66}         & \textbf{62.01}         & \textbf{7.18}          & \textbf{39.42}          & \textbf{21.17\sym{*}}         & \textbf{54.68\sym{*}}         \\ \midrule
SC2-7B  & 26.00                  & 61.68                  & 27.14                  & 56.52                  & 7.66                   & 39.54                   & 20.27                  & 52.58                  \\
\addlinespace[0.1em]
\hdashline
\addlinespace[0.3em]
\ \ \ + \ours      & \textbf{27.58}         & \textbf{63.84}         & \textbf{32.86}         & \textbf{64.07}         & \textbf{8.44}          & \textbf{41.03}          & \textbf{22.96\sym{*}}         & \textbf{56.31\sym{*}}         \\ \bottomrule
\end{tabular}

\vspace*{0.5\baselineskip}

\begin{tabular}{@{}lccccccS[table-format=2.3,table-align-text-post=false]S[table-format=2.3,table-align-text-post=false]@{}}
\toprule
                 & \multicolumn{6}{c}{\repoeval}                                                                                                                   & \multicolumn{2}{c}{\multirow{2}{*}{Average}}    \\ \cmidrule(r){2-7}
                 & \multicolumn{2}{c}{Line}                        & \multicolumn{2}{c}{API}                       & \multicolumn{2}{c}{Function}                    & \multicolumn{2}{c}{}                            \\ \midrule
                 & \multicolumn{1}{c}{EM} & \multicolumn{1}{c}{ES} & \multicolumn{1}{c}{EM} & \multicolumn{1}{c}{ES} & \multicolumn{1}{c}{EM} & \multicolumn{1}{c}{ES} & \multicolumn{1}{c}{EM\ \ \ \tiny{\ }} & \multicolumn{1}{c}{ES} \\ \midrule
DS-1.3B & 24.50                  & 50.42                  & 18.81                  & 58.15                  & 3.96                   & 29.73                   & 15.76                  & 46.10                  \\
\addlinespace[0.1em]
\hdashline
\addlinespace[0.3em]
\ \ \ + \ours      & \textbf{27.25}         & \textbf{53.45}         & \textbf{21.81}         & \textbf{59.79}         & \textbf{5.93}          & \textbf{31.92}          & \textbf{18.33\sym{*}}         & \textbf{48.39\sym{*}}         \\ \midrule
DS-6.7B & 26.62                  & 52.59                  & 22.69                  & 61.94                  & 7.47                   & 36.24                   & 18.93                  & 50.26                  \\
\addlinespace[0.1em]
\hdashline
\addlinespace[0.3em]
\ \ \ + \ours      & \textbf{30.31}         & \textbf{55.97}         & \textbf{25.12}         & \textbf{63.06}         & \textbf{7.69}          & \textbf{37.22}          & \textbf{21.04\sym{*}}         & \textbf{52.08\sym{*}}         \\ \midrule
SC2-3B  & 21.88                  & 46.74                  & 18.81                  & 56.66                  & 4.40                   & 29.99                   & 15.03                  & 44.46                  \\
\addlinespace[0.1em]
\hdashline
\addlinespace[0.3em]
\ \ \ + \ours       & \textbf{26.56}         & \textbf{50.56}         & \textbf{23.06}         & \textbf{61.02}         & \textbf{7.25}          & \textbf{33.79}          & \textbf{18.96\sym{*}}         & \textbf{48.46\sym{*}}         \\ \midrule
SC2-7B  & 27.94                  & 51.60                  & 21.56                  & 58.98                  & 6.81                   & 32.80                   & 18.77                  & 47.79                  \\
\addlinespace[0.1em]
\hdashline
\addlinespace[0.3em]
\ \ \ + \ours       & \textbf{34.19}         & \textbf{57.29}         & \textbf{27.31}         & \textbf{63.04}         & \textbf{8.35}          & \textbf{35.40}          & \textbf{23.28\sym{*}}         & \textbf{51.91\sym{*}}         \\ \bottomrule
\end{tabular}
\caption{\label{tab:cceval+repoeval}
Exact Match (EM) and Edit Similarity (ES) results of training w/o and w/ \ours for different code \llm{s} on \cceval \citep{crosscodeeval}, \ccleval \citep{repoformer}, and \repoeval \citep{repoeval} using greedy decoding, following the experimental setting of existing work \citep{repoformer}. Our evaluation is conducted in ``Retrieval'' mode, where evaluation prompts are constructed by prepending the retrieved cross-file context to the current file, to show the performance of repository-level cross-file code completion.
}
\end{table*}

We conduct controlled experiments for all the studied code \llm{s} in our experiments. Specifically, we conduct two continual pre-training experiments for each model as follows:
\begin{itemize}[leftmargin=1em]
    \item \textbf{\ntp}: existing pre-training scheme with \ntpfull (\ntp) objective only.
    \item \textbf{\ourstrain}: our newly proposed objective that incorporates \oursfull (\ours) objective with \ntpfull (\ntp) objective.
\end{itemize}

Throughout this section, we determine the end of generation solely based on \texttt{<eoi>} generated by the model without any rule-based \postp, unless otherwise specified (\S\ref{eval:reason}). We also conduct statistical analysis: results marked with \textbf{*} are statistically significant ($p<0.05$) based on paired t-tests.

\subsection{Syntax-Aware Multilingual Code \fim}\label{eval:safim}
We use \safim \citep{safim}, a syntax-aware and multilingual code \fimfull benchmark, to evaluate the effectiveness of \ours. \safim focuses on syntax-aware completions of program structures, covering algorithmic block (\ie Algo and Algo$_{\text{v2}}$), control-flow expression (\ie Control), and API function call (\ie API). It consists of 17,720 examples from four different programming languages, including Python, Java, C++, and C\#. \safim employs execution-based evaluation and reports \passat1 as the evaluation metric. As shown in Table \ref{tab:safim}, compared with \ntp only, adding \ours achieves up to 5\% improvements on average across all the studied code \llm{s}. 
Importantly, the improvement is consistent across languages and program structures,  demonstrating \ours's ability to enhance \fim capabilities regardless of language or completion context.

\subsection{Repository-Level Cross-File Code \fim}\label{eval:repo}

In addition to single-file \fim evaluation with \safim,  we also evaluate the effectiveness of \ours on repository-level code \fimfull in cross-file scenarios via \cceval \citep{crosscodeeval}, \ccleval \citep{repoformer}, and \repoeval \citep{repoeval}. \cceval (Python) and \ccleval are two repository-level cross-file benchmarks that leverage more than 1500 raw Python repositories to construct 12,665 examples across line, chunk, and function completion tasks, which are used for a more rigorous evaluation. \repoeval is another repository-level cross-file code completion benchmark consisting of 3,655 line, API, and function completion tasks created from 32 Python repositories. We follow existing work \citep{repoformer} to evaluate the model's \fim performance on these benchmarks and use Exact Match (EM) and Edit Similarity (ES) as our evaluation metrics. As shown in Table \ref{tab:cceval+repoeval}, adding \ours provides consistent improvements for all models across different benchmarks and completion tasks. Specifically, \ours achieves up to 24\% improvements on EM and 9\% improvements on ES relatively, showing its significant effectiveness.
\begin{table}[!ht]
\centering
\sisetup{table-space-text-post = \sym{*}}
\sisetup{table-number-alignment=center}
\resizebox{0.95\columnwidth}{!}{
\begin{tabular}{@{}lcSS@{}}
\toprule
                                                   & \textbf{Code Repair} & \multicolumn{2}{c}{\textbf{Code Reasoning}} \\ \cmidrule(l){2-2} \cmidrule(l){3-4}
                                                   & \dfourj         & \multicolumn{1}{c}{\crux-I\ \ \ \ \tiny{\ }}           & \multicolumn{1}{c}{\crux-O}           \\ \midrule
DS-1.3B                                            & 33                & 42.0                 & 31.0                 \\
\addlinespace[0.1em]
\hdashline
\addlinespace[0.3em]
\ \ \ + \ours & \textbf{39}       & \textbf{44.7\sym{*}}        & \textbf{31.8\sym{*}}        \\ \midrule
DS-6.7B                                            & 58                & 52.1                 & 39.2                 \\
\addlinespace[0.1em]
\hdashline
\addlinespace[0.3em]
\ \ \ + \ours & \textbf{59}       & \textbf{52.4}\ \ \tiny{\ }        & \textbf{39.6}\ \ \tiny{\ }        \\ \midrule
SC2-3B                                             & 39                & 42.8                 & 32.1                 \\
\addlinespace[0.1em]
\hdashline
\addlinespace[0.3em]
\ \ \ + \ours & \textbf{41}       & \textbf{43.9\sym{*}}        & \textbf{32.6\sym{*}}        \\ \midrule
SC2-7B                                             & 41                & 44.4                 & 35.9                 \\
\addlinespace[0.1em]
\hdashline
\addlinespace[0.3em]
\ \ \ + \ours & \textbf{47}       & \textbf{45.5\sym{*}}        & \textbf{36.1\sym{*}}        \\ \bottomrule
\end{tabular}
}
\caption{\label{tab:dfourj+cruxeval}
Code fixing and reasoning performance of models trained w/o and w/ \ours on \dfourj and \crux. On \dfourj, following the convention, we report the number of patches that passed test suites under greedy decoding. On \crux, we follow the original setting to do sampling with $T=0.2$ and $n=10$ and to extract accurate input/output values from raw generation.
}
\vspace*{-0.8\baselineskip}
\end{table}

\subsection{Code Repair via \fimfull}\label{eval:fix}
To assess \ours's impact on practical applications beyond code completion, we evaluate the performance of code repair using \dfourj \citep{defects}. \dfourj consists of open-source bugs found across 15 Java repositories. Following existing works \citep{fitrepair, chatrepair}, we collect 313 single-hunk bugs from \dfourj that can be fixed by replacing or adding a continuous code hunk. Specifically, for each bug, models are prompted to generate the correct code hunk (\ie patch) given the left and right contexts of the buggy code hunk, with correctness verified by project test suites. 
As shown in the ``Code Repair'' section of Table \ref{tab:dfourj+cruxeval}, adding \ours during training results in relatively up to 18\% more bugs fixed by the model\footnote{Note that \dfourj is a small dataset with only 313 examples and models can only solve 30-60 out of those, which makes it hard to obtain statistically significant differences with greedy decoding.}, suggesting that enhanced planning capabilities translate directly to better bug-fixing performance.

\subsection{Code Reasoning via \fimfull}\label{eval:reason}
We further examine whether \ours's planning benefits extend to code reasoning tasks. We use \crux \citep{cruxeval} which comprises 800 Python functions paired with two distinct tasks: \crux-I, where \llm{s} need to predict the input from the known output, and \crux-O, where \llm{s} are required to predict the output based on the given input. 

We reformat prompts of \crux-I into \fim style and leave \crux-O as \ltor generation, both of which are evaluated in zero-shot setting. Different from previous subsections where \postp is not used, we follow the same pipeline as in the original \crux paper to extract accurate input/output values from generation because we are focusing on evaluating the reasoning capability of \llm{s} rather than their capability of generating correct code\footnote{In \crux-I, we only want to evaluate the correctness of the input value infilled by \llm{s} in the given assertion. However, \fim-style prompts we use in the experiments does not restrict \llm{s} from writing multiple assertions before starting infilling the given assertion, which is useless in this task. So we use \postp techniques to extract the input value infilled for the given assertion to better evaluate the reasoning capabilities.}. 
As shown in the ``Code Reasoning'' section of Table \ref{tab:dfourj+cruxeval}, \ours demonstrates up to 6\% improvements on both \crux-I and \crux-O tasks for all the code \llm{s} consistently, which shows that \ours also improves intrinsic code reasoning capabilities of \llm{s}.

\section{Discussion}

\subsection{\ntp Alone Cannot Yield Horizon Awareness}\label{sec:hlppre}

A key question of interest is whether standard next-token prediction (\ntp) inherently provides models with awareness of generation horizons. Through systematic analysis, we demonstrate that hidden states of models trained with \ntp alone do not capture meaningful information about required completion lengths.

\begin{figure*}[!ht]
  \centering
  \subfloat[DS-Coder 1.3B]
  {\includegraphics[width=0.252\textwidth]{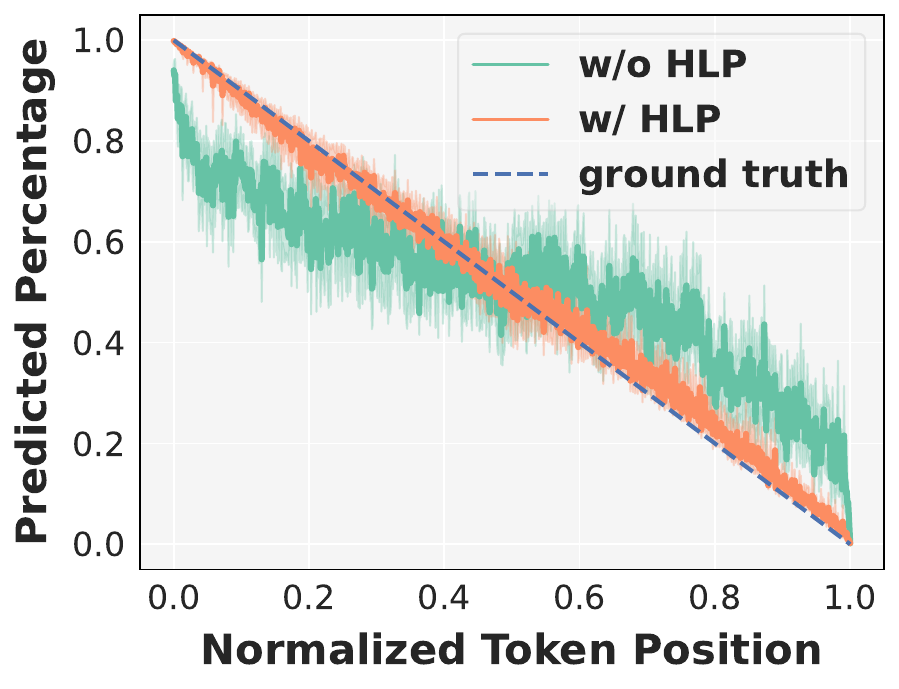}\label{fig:probing-line-ds-1b}}
  \hspace{1mm}
  \subfloat[DS-Coder 6.7B]
  {\includegraphics[width=0.235\textwidth]{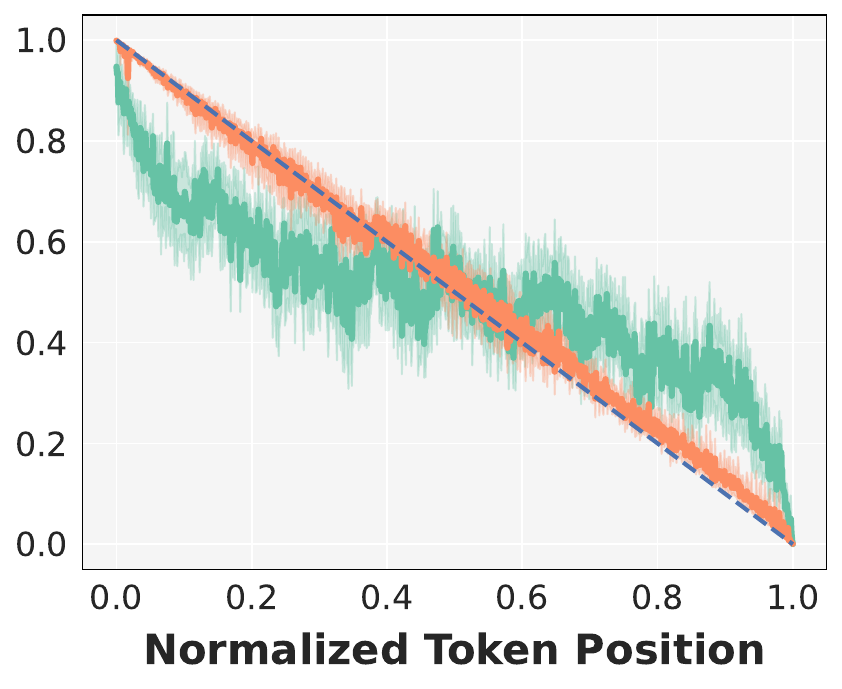}\label{fig:probing-line-ds-7b}}
  \hspace{1mm}
  \subfloat[\starcodertwo 3B]
  {\includegraphics[width=0.235\textwidth]{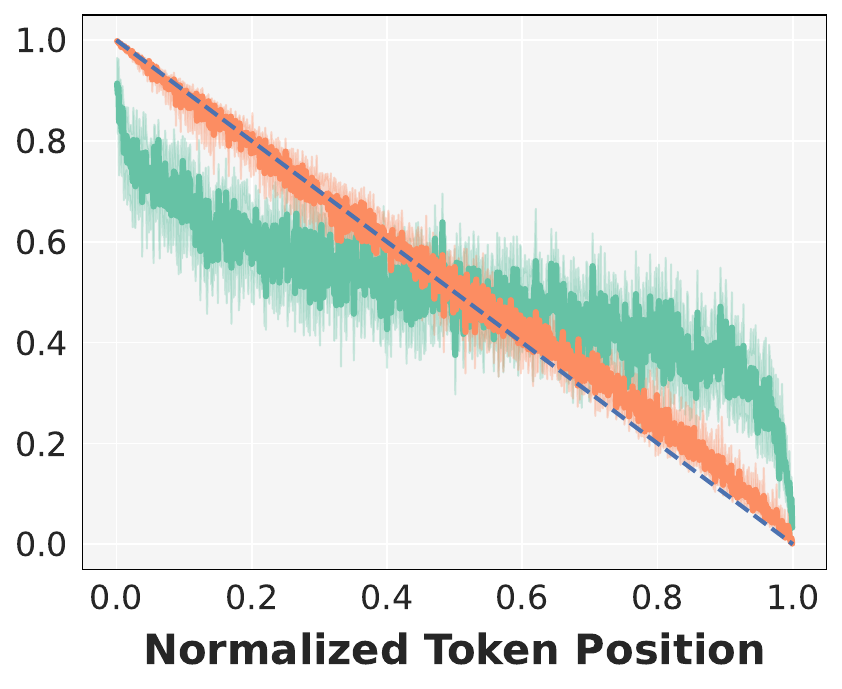}\label{fig:probing-line-sc2-3b}}
  \hspace{1mm}
  \subfloat[\starcodertwo 7B]
  {\includegraphics[width=0.235\textwidth]{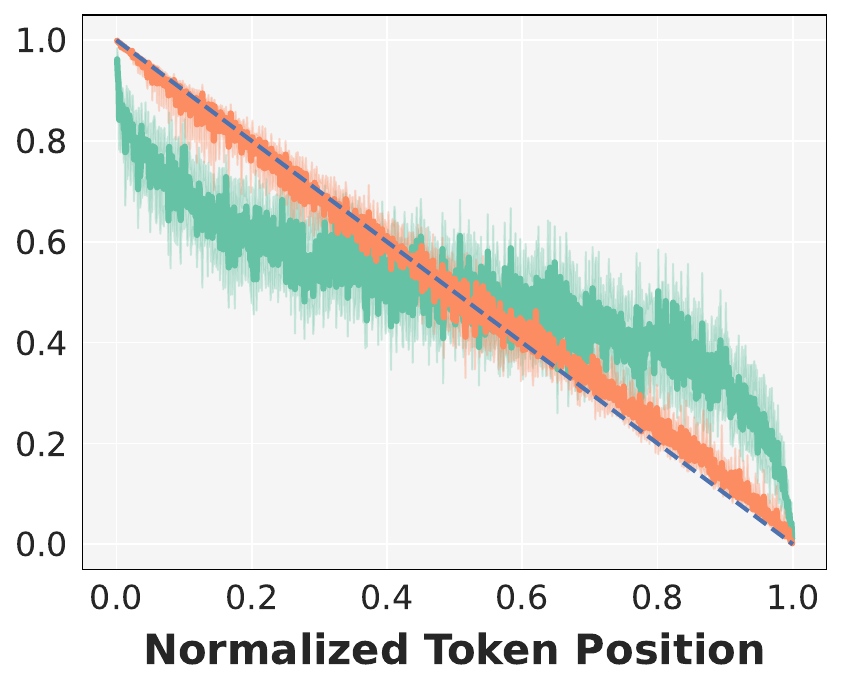}\label{fig:probing-line-sc2-7b}}
  \caption{Predicted percentage of remaining future tokens (as defined in Eq. (\ref{formula:ftc_label})) from models trained w/o and w/ \ours at different token positions, where the position of each token is normalized to the corresponding percentage over the sequence.}
  \label{fig:probing_line}
\vspace*{-\baselineskip}
\end{figure*}

To quantify this, we design a probing task that attempts to predict the remaining completion length from models' hidden states. We extract hidden states from 20K code snippets (approximately 7.8M tokens) using models trained with and without \ours and split the data to ensure no sequence-level overlap between train and test. Using these hidden states as input and normalized remaining token counts as targets, we fit linear regression models while keeping the underlying transformer parameters frozen.

\begin{table}[h]
\centering
\resizebox{0.9\columnwidth}{!}{
\begin{tabular}{@{}lcccc@{}}
\toprule
Test $\uparrow$   & DS-1.3B        & DS-6.7B        & SC2-3B         & SC2-7B         \\ \midrule
NTP     & 0.440          & 0.519          & 0.356          & 0.410          \\
\addlinespace[0.1em]
\hdashline
\addlinespace[0.3em]
NTP+HLP & \textbf{0.915} & \textbf{0.913} & \textbf{0.932} & \textbf{0.932} \\ \bottomrule
\end{tabular}
}
\captionof{table}{Probing results of models trained w/o and w/ \ours. We report the coefficient of determination $(R^2)$ of prediction, which is the higher the better.}
\label{tab:probing}
\end{table}

Figure \ref{fig:probing_line} shows predicted versus actual remaining token percentages at different positions, while Table \ref{tab:probing} reports the coefficient of determination $(R^2)$. Models trained with \ntp alone show poor fit, indicating their hidden states lack horizon information. In contrast, models trained with \ours show much stronger correlations between hidden states and remaining lengths. This result demonstrates that horizon awareness must be explicitly trained rather than emerging naturally from \ntp. We also carried out a non-linear probing experiment with the same inputs and targets by replacing linear regression models with an MLP regressor that uses logistic activation function (Appendix \ref{sec:nonlinearprobing}), which further confirms our finding here.

\subsection{Why \OURSfull works?}\label{sec:analysis}

The effectiveness of \OURSfull (\ours) stems from its ability to address a critical limitation of standard next-token prediction (\ntp) in autoregressive models: the absence of global planning awareness. While \ntp optimizes for local token-level coherence by maximizing the likelihood of immediate next tokens, it inherently struggles to enforce long-horizon structural consistency between the generated \fimmid and the given \fimsuf in \fim. This limitation becomes pronounced when the model must align generated code with both preceding logic and subsequent constraints (e.g., API calls or variable dependencies defined in the suffix). 

\ours bridges this gap by teaching models to explicitly condition the model on the remaining generation horizon at every decoding step. By regressing the normalized count of future tokens required to complete \fimmid, \ours effectively decomposes the infilling task into a sequence of length-aware subgoals, where each generated token is tied to a dynamically updated ``budget'' of remaining steps. This mechanism mirrors human problem-solving strategies, where progress estimation (e.g., "30\% of steps remaining") guides iterative refinement of intermediate actions to meet global objectives.

\begin{figure}[b]
\vspace*{-\baselineskip}
\centering
\includegraphics[width=0.88\linewidth]{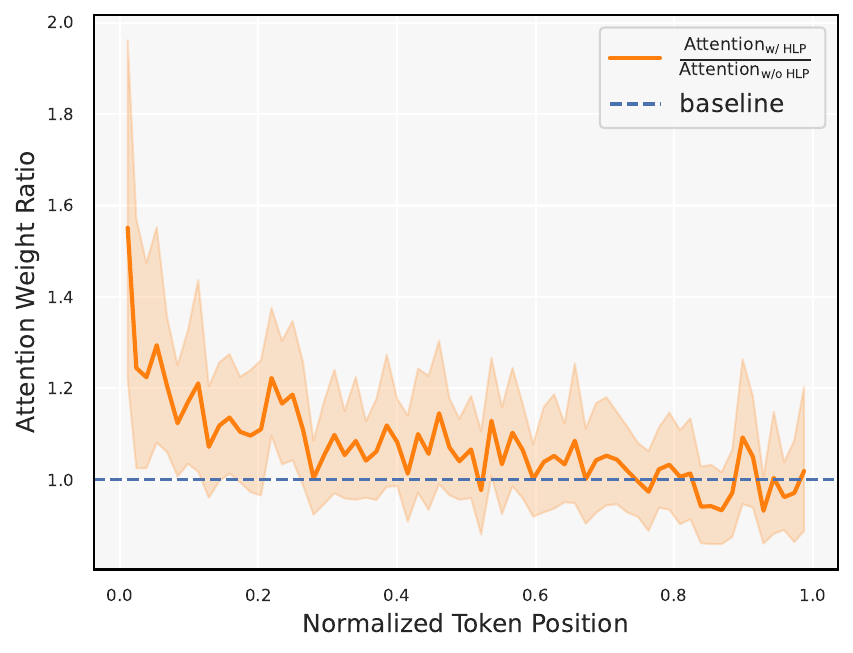}
\caption{Attention analysis of \dscoderbase 1.3B on \safim, showing the ratio of attention paid to \fimsuf between models trained with and without \ours. The X-axis shows the normalized position in the sequence, and the Y-axis shows the attention ratio. Values above 1 indicate that the \ours model pays more attention to \fimsuf than the baseline model. We observe that the model trained with \ours generally pays more attention to \fimsuf, especially at the beginning, demonstrating its lookahead planning behavior.}
\label{fig:attention_analysis}
\vspace*{-0.3\baselineskip}
\end{figure}

The success of \ours is rooted in its complementary role to \ntp. While \ntp ensures local fluency, \ours provides a global scaffold for planning. By unifying local token prediction with global horizon awareness, \ours bridges the gap between autoregressive decoding and holistic reasoning, enabling models to dynamically adapt their generation strategy to long-horizon constraints. This synergy is particularly critical in code infilling, where the correctness of \fimmid depends on both preceding logic (\ie \fimpre) and subsequent context (\ie \fimsuf), demanding a balance between immediate token likelihood followed by \fimpre and forward-looking structural coherence constrained by distant \fimsuf. 

Our in-depth attention analysis provides direct evidence of this enhanced planning capability. Figure \ref{fig:attention_analysis} shows that models trained with \ours pay significantly more attention to \fimsuf context, particularly at the start of the generation. This indicates that \ours enables proactive planning by considering long-horizon constraints from the beginning. Furthermore, the consistent improvements across \fim, bug-fixing, and reasoning tasks demonstrate that this enhanced planning capability generalizes beyond simple completion scenarios.

In summary, \ours transforms code generation from a myopic token-by-token process into a structured planning task. It enables models to dynamically adapt their generation strategy to long-horizon constraints, ensuring that each step not only satisfies local fluency but also contributes to a globally coherent solution. This dual focus on immediate actions and overarching goals positions \ours as a pivotal advancement in code LLMs.

\subsection{\ours Mitigates Planning Failures in \fim}\label{sec:quantitative}

We conduct a quantitative analysis to empirically demonstrate the frequency of planning failures in \fim and evaluate the effectiveness of \ours in addressing these failures. As presented in Figure \ref{fig:fim_plan_example} (\S\ref{sec:postp}), a prevalent manifestation of planning failures in \fim is the "correct code but incorrect order" pattern, wherein the model generates the final line of ground truth code accurately but positions it incorrectly within the completion. To maintain analytical tractability, our study focuses specifically on planning failures that result in this particular pattern. We establish the following criteria to identify instances of the "correct code but incorrect order" pattern in model's \fim completions:
\begin{itemize}[leftmargin=1em]
    \item Completion fails to pass all the unit tests.
    \item Completion contains the last line from ground truth.
    \item Completion does not end with this last line.
\end{itemize}

Testing on \safim's Algorithmic and Algorithmic\_v2 partitions which focus on multi-line \fim, we find that models trained without \ours exhibit the “correct code but incorrect order” pattern in 5-10\% of problems across the benchmark, as shown in Table \ref{tab:quantitative_analysis}. With \ours training, this drops to 4-6\%, demonstrating that \ours helps alleviate lack-of-planning failures effectively.

\begin{table}[ht]
\centering
\begin{tabular}{@{}lccc@{}}
\toprule
                & \multicolumn{2}{c}{SAFIM}                                                                                                    & \multicolumn{1}{c}{\multirow{2}{*}{Avg}} \\ \cmidrule(r){2-3}
                & \multicolumn{1}{c}{Algo} & \multicolumn{1}{c}{Algo$_{\text{v2}}$}  & \multicolumn{1}{c}{}                         \\ \midrule
DS-1.3B    & 9.34\%                            & 10.48\%                             & 9.91\%                                         \\
\addlinespace[0.1em]
\hdashline
\addlinespace[0.3em]
\ \ \ + \ours & \textbf{6.83\%}                   & \textbf{6.52\%}                  & \textbf{6.68\%}           \\ \midrule
DS-6.7B    & 5.16\%                            & 5.93\%                             & 5.55\%                                         \\
\addlinespace[0.1em]
\hdashline
\addlinespace[0.3em]
\ \ \ + \ours & \textbf{4.36\%}                   & \textbf{4.44\%}                  & \textbf{4.40\%}           \\ \midrule
SC2-3B    & 7.25\%                            & 7.35\%                             & 7.30\%                                         \\
\addlinespace[0.1em]
\hdashline
\addlinespace[0.3em]
\ \ \ + \ours & \textbf{5.65\%}                   & \textbf{5.43\%}                  & \textbf{5.54\%}           \\ \midrule
SC2-7B    & 5.79\%                            & 6.26\%                             & 6.03\%                                         \\
\addlinespace[0.1em]
\hdashline
\addlinespace[0.3em]
\ \ \ + \ours & \textbf{4.69\%}                   & \textbf{4.29\%}                  & \textbf{4.49\%}           \\
\bottomrule
\end{tabular}
\caption{\label{tab:quantitative_analysis}
Quantitative analysis on the frequency of lack-of-planning issue in models trained w/ and w/o \ours on \safim. We follow the same settings used in  \S\ref{eval:safim}.
}
\end{table}

\section{Related Work}\label{sec:related}

\textbf{\fimfull for Code Language Models }
Large language models trained on massive source code data have demonstrated great potential in various applications for software development. While early models such as Codex \citep{humaneval} and CodeGen \citep{codegen} only support Left-to-Right (L2R) generation, Fill-in-the-Middle (or infilling) has attracted increased attention because right context naturally carries an indispensable part of information for completing code in the middle \citep{incoder, openaifim}. Subsequently, FIM training has become a common practice widely adopted by most code LLMs, such as StarCoder \citep{starcoder, starcodertwo}, DeepSeek-Coder \citep{dscoder, dscodervtwo}, and Code Llama \citep{codellama}. 

Existing models generally tackle the infilling problem by breaking a code snippet into \fimpre-\fimmid-\fimsuf, and reordering them into \fimpre-\fimsuf-\fimmid (PSM) or \fimsuf-\fimpre-\fimmid (SPM), which are then used for next-token prediction (\ntp) training. We point out that the infilling task cannot be effectively learned with \ntp alone, as it requires planning capability for the model to fluently and meaningfully connect the middle completion to \fimsuf through forward looking during auto-regressive decoding.

An alternative approach is to train two language models in different directions, with one generating from left to right and the other from right to left, and have the two generations meet in the middle \citep{meet-in-the-middle}. Nevertheless, the L2R model does not have access to the right context, and vice versa, which impedes holistic planning that takes into account the context from both sides.

\textbf{Planning and Lookahead in Language Generation } Standard decoder-only models are trained with next-token prediction and used to sequentially predict one token at a time, conditioned only on past tokens, in an auto-regressive manner. One drawback of this paradigm is that models are not aware of future tokens during decoding. The token that maximizes the conditional probability at current step may lead to suboptimal continuation, and consequently the model can fail to compose a fluent and sensible generation that meets human requirements. Various decoding techniques have been proposed to address the problem through tree search with lookahead heuristics, particularly for constrained generation problems \citep{NeuroLogic, DeAL}. While these methods are training-free, they inevitably incur additional cost of inference complexity. 

Apart from those, \citet{multitoken} proposed to predict multiple tokens from a single hidden state during both training and inference, which was shown to achieve stronger performance on coding tasks with no computation overhead. While multi-token prediction enhances models' planning capability within the $n$ tokens predicted together ($n\leq 8$), we argue that with a small $n$, the limited horizon is usually insufficient for planning in the case of infilling as the connection from \fimmid to \fimsuf only happens towards the end of the generation. In contrast, \ours adopts a global and arbitrary long horizon over all future tokens by counting the remaining generation budget, which more effectively helps models to close the generation fluently with early planning\footnote{Please refer to Appendix \S\ref{sec:hlpmtp} for more details.}.

\section{Conclusion and Future Work}

\fimfull is ubiquitous in code completion, reflecting the iterative nature of software development where code is frequently inserted between existing sections, and thus has become an important consideration in the development of code language models. The current \fim training paradigm splits and reorders training sequences for next-token prediction. However, this approach frequently results in models struggling to generate content that smoothly aligns with the right context. While existing \fim benchmarks rely on different post-processing methods to circumvent this problem, we emphasize that such methods typically require dataset-specific assumptions, which are impractical in real-world scenarios.

To address this limitation and enhance the infilling capability of code language models, we propose \OURSfull (\ours). \ours teaches models to predict the portion of remaining tokens at every step,  enabling them to develop planning awareness over arbitrarily long horizons. Experiments across different model families and sizes show that \ours improves infilling performance on diverse \fim benchmarks, across file-level and repository-level, and without using any dataset-specific post-processing.  Moreover, the enhanced planning capability acquired through \ours training also boosts models' performance on code reasoning tasks, suggesting broader benefits for language models' reasoning capabilities. Importantly, \ours achieves these improvements while maintaining efficiency, with negligible training overhead and no additional inference cost.

Our work marks a significant advancement in developing more effective code language models for real-world applications. Future work could explore extending \ours to other generation tasks requiring long-horizon planning and investigate its potential for enhancing general reasoning capabilities.

\section*{Limitations}

While \ours has proven effective through extensive evaluation in the paper, our pre-training experiments are restricted to models of $\leq$7B parameters due to the computation budget. It is prohibitively expensive to perform pretraining experiments for \llm, and unfortunately we do not have enough resources to demonstrate the impact of \ours on larger models. This highlights a broader challenge in open science for pre-training research. 
In addition, this work has mainly focused on the code domain, but the idea of horizon-length prediction can be broadly applicable for improving models' reasoning capability in natural language and mathematical domains as well, which we leave to future work.

\bibliography{custom}

\appendix
\section{Appendix}

\subsection{Effect of $\lambda$ Value Selection}\label{sec:lambdastudy}

Following the original experimental settings (but with 100k training steps given limited computational resources), we studied a series of $\lambda$ values and report both \ours validation loss and \safim pass@1. As shown in Table \ref{tab:lambdavalue}, our method is robust to hyperparameter selection: all non-zero $\lambda$ values improve over the baseline ($\lambda=0$). Moreover, we observe that \ours loss quickly decreases within the first few thousand training steps with all positive $\lambda$’s, indicating that a large $\lambda$ isn’t necessary for models to learn \ours well. This also coincides with our observation on downstream performance: while lower \ours loss generally correlates with better \safim performance, the improvement plateaus beyond a certain threshold, suggesting that once the model acquires basic horizon awareness, further optimization of \ours loss provides diminishing returns for downstream performance.

\begin{table}[h]
\centering
\begin{tabular}{@{}ccc@{}}
\toprule
$\lambda$ & \ours Valid Loss & \safim (pass@1)     \\ \midrule
0         & 0.254                           & 47.40\%          \\
\midrule
0.01      & 0.060                           & 48.98\%          \\
0.1       & 0.048                           & 49.10\%          \\
0.2       & 0.045                           & 48.74\%          \\
0.5       & \textbf{0.043}                  & \textbf{49.34\%} \\ \bottomrule
\end{tabular}
\captionof{table}{Study over the effect of $\lambda$ value selection on \ours valudation loss and \safim performance.}
\label{tab:lambdavalue}
\end{table}

\subsection{Non-Linear Probing for Horizon Awareness}\label{sec:nonlinearprobing}

By replacing the linear regression models in the original linear probing experiment with an MLP regressor that uses logistic activation function, we further conduct a non-linear probing experiment to study whether \ntp is able to yield horizon awareness. As shown in Table \ref{tab:nonlinearprobing}, even under the assumption of the non-linear relationship, models trained without \ours cannot effectively extract horizon information from their hidden states. This demonstrates that horizon awareness must be explicitly trained rather than emerging naturally from \ntp.

\begin{table}[h]
\centering
\resizebox{0.9\columnwidth}{!}{
\begin{tabular}{@{}lcccc@{}}
\toprule
Test $\uparrow$   & DS-1.3B        & DS-6.7B        & SC2-3B         & SC2-7B         \\ \midrule
NTP     & 0.392          & 0.431          & 0.180          & 0.226          \\
\addlinespace[0.1em]
\hdashline
\addlinespace[0.3em]
NTP+HLP & \textbf{0.897} & \textbf{0.885} & \textbf{0.856} & \textbf{0.871} \\ \bottomrule
\end{tabular}
}
\captionof{table}{Non-linear probing results of models trained w/o and w/ \ours. We report the coefficient of determination $(R^2)$ of prediction, which is the higher the better.}
\label{tab:nonlinearprobing}
\end{table}

\subsection{Comparing \ours with Multi-token Prediction}\label{sec:hlpmtp}

\begin{table*}[ht]
\centering
\begin{tabular}{@{}lccccc@{}}
\toprule
                & \multicolumn{4}{c}{SAFIM}                                                                                                    & \multicolumn{1}{c}{\multirow{2}{*}{Average}} \\ \cmidrule(r){2-5}
                & \multicolumn{1}{c}{Algo} & \multicolumn{1}{c}{Algo$_{\text{v2}}$} & \multicolumn{1}{c}{Control} & \multicolumn{1}{c}{API} & \multicolumn{1}{c}{}                         \\ \midrule
DS-1.3B    & 39.8                            & 42.4                               & 52.4                        & 56.1                     & 47.7                                         \\
\addlinespace[0.1em]
\hdashline
\addlinespace[0.3em]
\ \ \ + \ours & \textbf{41.3}                   & \textbf{46.1}                      & \textbf{53.4}               & \textbf{59.0}            & \textbf{50.0}                                \\
\addlinespace[0.1em]
\hdashline
\addlinespace[0.3em]
\ \ \ + multi-token prediction & 38.1                   & 41.3                      & 51.7               & 55.8            & 46.7                                \\ \bottomrule
\end{tabular}
\caption{\label{tab:discussion_mtp}
\Passat1 results of training w/ \ours and w/ multi-token prediction for \dscoderbase 1.3B on \safim \citep{safim} computed with greedy decoding.
}
\end{table*}

As discussed in \S\ref{sec:related}, we argue that multi-token prediction~\citep{multitoken} is insufficient for planning in \fim, because multi-token prediction only enhances models’ planning capability for a short and limited horizon, which does not suites \fim well as the connection from \fimmid to \fimsuf happens over a long horizon. Instead, \ours focuses on long-horizon planning and is more effective for \fim. We conduct an experiment following the same settings in \S\ref{sec:setup} to compare \ours with multi-token prediction on \dscoderbase 1.3B by predicting the next 4 tokens~\citep{multitoken}. We report their performance on \safim. As shown in Table \ref{tab:discussion_mtp}, while adding \ours to \ntp largely improves the model’s performance on \safim, multi-token prediction fails to do so. These results provide empirical evidence that long-horizon planning capabilities brought by \ours is essential for advancing \fim performance.

\subsection{\ours Improves Self-Infilling Performance}\label{sec:selfinfilling}
Recently, some strategies have been proposed to conduct \ltor code generation based on \fim capabilities and self-infilling~\citep{zheng2024selfinfillingcodegeneration} is one of the representative strategies of this kind. Given an initial input prompt \fimpre, self-infilling performs \ltor code generation by using \fim decoding to (1) generate \fimsuf based on \fimpre and then (2) generate \fimmid based on \fimpre and previously generated \fimsuf. Furthermore, self-infilling proposes a looping mechanism to improve the generated code iteratively, where it first uses \ltor decoding to generate a new \fimsuf$'$ based on \fimpre and previously generated \fimmid and then uses \fim decoding to generate a new \fimmid$'$ based on \fimpre and this newly generated \fimsuf$'$. This looping procedure can be continued for multiple rounds to obtain greater improvements.

\begin{table}[h]
\centering
\begin{tabular}{@{}lccccc@{}}
\toprule
\humaneval   & $N=0$        & $1$        & $2$         & $3$     & $4$     \\ \midrule
$\text{DS-1.3B}$     & 27.4          & 29.3          & 31.7          & 31.7  & 31.7        \\
\addlinespace[0.1em]
\hdashline
\addlinespace[0.3em]
\ \ \ + \ours & \textbf{30.5} & \textbf{31.7} & \textbf{31.7} & \textbf{32.3} & \textbf{32.9} \\ \bottomrule
\end{tabular}
\captionof{table}{Self-infilling \humaneval performance of models trained w/o and w/ \ours, with $N$ ranging from 0 to 4. $N$ denotes the number of times the decoding process goes through the loop and $N = 0$ represents that the looping mechanism is not activated. We report \passat1 results using greedy decoding.}
\label{tab:selfinfilling}
\end{table}

Since self-infilling relies on \fim capabilities for \ltor code generation, it is interesting to study whether \ours can also enhance its performance. To this end, we evaluate the self-infilling performance of \dscoderbase 1.3B trained with and without \ours on \humaneval using greedy decoding, following the setting of the original paper~\citep{zheng2024selfinfillingcodegeneration}. We report models’ performance on \humaneval with $N$ ranging from 0 to 4, where $N$ denotes the number of times the decoding process goes through the loop and $N = 0$ represents that the looping mechanism is not activated. As shown in Table \ref{tab:selfinfilling}, the model trained with \ours consistently outperforms the model trained without \ours across different number of self-improving steps. Furthermore, with more self-improving steps, while the performance of the model trained without \ours gets stuck, the performance of the model trained with \ours continues to showcase steady improvements.

\subsection{Effect of \ours on \ltorfull Performance}\label{sec:hlpltor}
While \ours have significantly improved the \fim performance of \llm{s}, we also study its impact on the \ltor code completion. 
To this end, we evaluate \ltor performance on \humaneval \citep{humaneval} and \mbpp \citep{mbpp} with \dscoderbase 1.3B. We further employ \humanevalp and \mbppp from \evalplus~\cite{evalplus} for more rigorous evaluation with better test coverage. 
As shown in Table \ref{tab:discussion_l2r}, with \ours applied to \fim data only (\ie $\text{\ours}_{\text{\fim}}$), the performance on \ltor tasks sometimes shows a slight degradation. We hypothesize that applying \ours to \fimmid only causes unbalanced training on \fimpre and \fimsuf parts.

\begin{table*}[ht]
\centering
\begin{tabular}{@{}lccc@{}}
\toprule
                                                                & \multicolumn{2}{c}{\ltorfull}                                                                & \multicolumn{1}{c}{\fimfull} \\ \midrule
                                                                & \multicolumn{1}{c}{\humaneval (+)} & \multicolumn{1}{c}{\mbpp (+)} & \multicolumn{1}{c}{\safim}              \\ \midrule
$\text{DS-1.3B}$                                                   & \textbf{26.3 (22.0)}                              & \textbf{45.8 (36.7)}                          & 47.7                                   \\
\addlinespace[0.1em]
\hdashline
\addlinespace[0.3em]
\ \ \ + \ours$_{\text{\fim}}$                               & 25.5 (21.3)                                       & 45.8 (36.5)                                   & \textbf{50.0}                          \\ \midrule
\ \ \ + \ours$_{\text{\fim}}$ + \ours$_{\text{\ltor}}$ & \textbf{26.2 (22.0)}                              & \textbf{45.7 (36.6)}                          & \textbf{49.6}                          \\ \bottomrule
\end{tabular}
\caption{\label{tab:discussion_l2r}
Effect of $\text{\ours}_{\text{\fim}}$ only and $\text{\ours}_{\text{\fim}} + \text{\ours}_{\text{\ltor}}$ for \dscoderbase 1.3B on \ltor and \fim tasks. On \ltor tasks including \humaneval (+) and \mbpp (+), we do sampling with $T=0.8$ and $n=200$. We report \passat1 performance of all the models, where numbers outside and inside parenthesis ``()'' indicate \texttt{base} and \texttt{plus} versions of \evalplus, respectively. For \fim experiments on \safim, we follow the same settings used in \S\ref{eval:safim}.
}
\end{table*}

\begin{table*}[ht]
\centering
\begin{tabular}{@{}lccccc@{}}
\toprule
                & \multicolumn{4}{c}{SAFIM}                                                                                                    & \multicolumn{1}{c}{\multirow{2}{*}{Average}} \\ \cmidrule(r){2-5}
                & \multicolumn{1}{c}{Algo} & \multicolumn{1}{c}{Algo$_{\text{v2}}$} & \multicolumn{1}{c}{Control} & \multicolumn{1}{c}{API} & \multicolumn{1}{c}{}                         \\ \midrule
DS-1.3B    & 39.8                            & 42.4                               & 52.4                        & 56.1                     & 47.7                                         \\
\addlinespace[0.1em]
\hdashline
\addlinespace[0.3em]
\ \ \ + \ours (linear) & 41.3                   & \textbf{46.1}                      & 53.4               & \textbf{59.0}            & \textbf{50.0}                                \\
\addlinespace[0.1em]
\hdashline
\addlinespace[0.3em]
\ \ \ + \ours (mlp) & \textbf{41.6}                   & 45.9                      & \textbf{54.0}               & 57.4            & 49.7                                \\ \bottomrule
\end{tabular}
\caption{\label{tab:ablation_hlp}
\Passat1 results of \ours w/ linear layer as $\mathsf{hlp\_head}$ (\ie ``\ours (linear)'') and w/ MLP layer as $\mathsf{hlp\_head}$ (\ie ``\ours (mlp)'') for \dscoderbase 1.3B on \safim \citep{safim} computed with greedy decoding.
}
\end{table*}

\begin{table*}[ht]
\centering
\sisetup{table-space-text-post = \sym{*}}
\sisetup{table-number-alignment=center}
\begin{tabular}{@{}lccccc@{}}
\toprule
                & \multicolumn{4}{c}{SAFIM}                                                                                                    & \multicolumn{1}{c}{\multirow{2}{*}{Average}} \\ \cmidrule(r){2-5}
                & \multicolumn{1}{c}{Algo} & \multicolumn{1}{c}{Algo$_{\text{v2}}$} & \multicolumn{1}{c}{Control} & \multicolumn{1}{c}{API} & \multicolumn{1}{c}{}                         \\ \midrule
DS-1.3B    & 39.8                            & 42.4                               & 52.4                        & 56.1                     & 47.7                                         \\
\addlinespace[0.1em]
\hdashline
\addlinespace[0.3em]
\ \ \ + \ours (all) & \textbf{41.3}                   & \textbf{46.1}                      & \textbf{53.4}               & \textbf{59.0}            & \textbf{50.0}                                \\
\addlinespace[0.1em]
\hdashline
\addlinespace[0.3em]
\ \ \ + \ours (first) & 40.0                   & 44.4                      & 52.3               & 57.4            & 48.5                                \\ \bottomrule
\end{tabular}
\caption{\label{tab:ablation_first_token_only}
\Passat1 results of applying \ours to all tokens (\ie “\ours (all)”) and first token only (\ie “\ours (first)”) for \dscoderbase 1.3B on \safim \citep{safim} computed with greedy decoding.
}
\end{table*}

\begin{table*}[h]
\centering
\sisetup{table-space-text-post = \sym{*}}
\sisetup{table-number-alignment=center}
\begin{tabular}{@{}lccccc@{}}
\toprule
                & \multicolumn{4}{c}{SAFIM}                                                                                                    & \multicolumn{1}{c}{\multirow{2}{*}{Average}} \\ \cmidrule(r){2-5}
                & \multicolumn{1}{c}{Algo} & \multicolumn{1}{c}{Algo$_{\text{v2}}$} & \multicolumn{1}{c}{Control} & \multicolumn{1}{c}{API} & \multicolumn{1}{c}{}                         \\ \midrule
DS-1.3B    & 39.8                            & 42.4                               & 52.4                        & 56.1                     & 47.7                                         \\
\addlinespace[0.1em]
\hdashline
\addlinespace[0.3em]
\ \ \ + \ours (normalized) & \textbf{41.3}                   & \textbf{46.1}                      & \textbf{53.4}               & \textbf{59.0}            & \textbf{50.0}                                \\
\addlinespace[0.1em]
\hdashline
\addlinespace[0.3em]
\ \ \ + \ours (raw) & 27.7                   & 29.8                      & 34.4               & 47.7            & 34.9                                \\ \bottomrule
\end{tabular}
\caption{\label{tab:ablation_normalization}
\Passat1 results of using normalized targets (\ie “\ours (normalized)”) and unnormalized targets (\ie “\ours (raw)”) in \ours for \dscoderbase 1.3B on \safim \citep{safim} computed with greedy decoding.
}
\end{table*}

To mitigate such effect, we need to devise another \ours task that can be applied to \ltor training (\ie $\text{\ours}_{\text{\ltor}}$). However, the original design of \ours in \S\ref{sec:ours_design} is not directly applicable to \ltor data. While the end of \fimmid in \fim data is strictly bounded by the beginning of \fimsuf, the end of \ltor data does not have any clear signals, as it is often possible to add additional contents (\eg another line of code or a new helper function) to the end of document fluently without any restrictions. 

Therefore, instead of taking the entire code file as the prediction horizon, we ask the model to predict the number of future tokens \textbf{required to complete current line} in \ltor training, which is a natural semantic unit in code. Furthermore, to avoid conflicts between $\text{\ours}_{\text{\fim}}$ and $\text{\ours}_{\text{\ltor}}$, we use two independent $\mathsf{hlp\_head}$s to let the model learn $\text{\ours}_{\text{\fim}}$ and $\text{\ours}_{\text{\ltor}}$ separately. As shown in Table \ref{tab:discussion_l2r}, by applying $\text{\ours}_{\text{\fim}}$ and $\text{\ours}_{\text{\ltor}}$ simultaneously, the performance degradation on \ltor tasks is recovered, with the improvement on \fim tasks largely retained. These results demonstrate the generalizable effectiveness of \ours and shows the huge potential of applying the idea of \ours to more general training scenarios.

\subsection{Additional Ablation Studies}\label{sec:ablation}

We conduct several ablation studies to justify the design choices of \ours. In this section, we conduct experiments using \dscoderbase 1.3B, follow the same settings in \S\ref{sec:setup}, and report models' performance on \safim.

\textbf{Complexity of $\mathsf{hlp\_head}$.} We conduct an experiment to study the effect of the complexity of $\mathsf{hlp\_head}$ by replacing the original linear layer (\ie ``\ours (linear)'') with a two-layer MLP with ReLU (\ie “\ours (mlp)”). As shown in Table \ref{tab:ablation_hlp}, increasing the complexity of $\mathsf{hlp\_head}$ does not bring significant improvements. We have also conducted a paired t-test between ``\ours (linear)'' and ``\ours (mlp)'' and did not see any clear directional statistical significance between them. Such results indicate that the complexity of $\mathsf{hlp\_head}$ does not have a major impact on performance.

\textbf{Applying \ours to all tokens \textit{v.s.} first token only.} While it is easy to see that knowing the \ours loss on the first token is sufficient to infer the horizon length in theory, having \ours loss on every token provides denser and more consistent supervision signals which makes learning easier (as discussed in \S\ref{sec:ours_design}). It also helps regularize the hidden representation of every subsequent token. To empirically show this, we conduct an experiment by applying \ours to the first token only (i.e., ``\ours (first)'') and compared its performance with our original \ours design (i.e., ``\ours (all)''). As shown in Table \ref{tab:ablation_first_token_only}, while applying \ours only to the first token performs better than \ntp only, applying \ours loss for each token can achieve better performance than applying it to just the first token.

\textbf{Normalized \textit{v.s.} unnormalized targets.} We use normalized targets in our original \ours design (\ie using $\frac{M-t}{M}$ rather than $M-t$) is that the scale of \ours loss will be otherwise in a huge range, e.g., some examples has single digit loss while some might have thousands. To further study the effect of normalization, we conduct an experiment by using $M-t$ as the target (\ie ``\ours (raw)'') rather than $\frac{M-t}{M}$ (\ie ``\ours (normalized)''). To achieve this, we remove the sigmoid function from the original \ours. As shown in Table \ref{tab:ablation_normalization}, setting the target as $M-t$ fails to improve \fim performance, likely due to the large-scale \ours loss after using $M-t$ as the target interferes with \ntp pre-training.

\end{document}